\documentclass{article}
\PassOptionsToPackage{table}{xcolor}
\usepackage{iclr2027_conference,times}

\usepackage[utf8]{inputenc}
\usepackage[T1]{fontenc}
\usepackage{hyperref}
\usepackage{url}
\usepackage{booktabs}
\usepackage{amsfonts}
\usepackage{amsmath}
\usepackage{amssymb}
\usepackage{nicefrac}
\usepackage{microtype}
\usepackage{xcolor}
\usepackage{graphicx}
\usepackage{array}
\usepackage{adjustbox}
\usepackage{makecell}
\usepackage{wrapfig}
\usepackage{algorithm}
\usepackage{algorithmic}

\newcommand{\method}{One-Forcing}
\newcommand{\softplus}{\operatorname{softplus}}
\newcommand{\stopgrad}{\operatorname{sg}}
\newcommand{\pdata}{p_{\mathrm{data}}}
\newcommand{\pgen}{p_{\theta}}

\newif\ifshowchanges
\showchangesfalse
\DeclareRobustCommand{\changedtext}[1]{{\ifshowchanges\color{red}\fi#1}}
\newcommand{\changedcolor}{\ifshowchanges\color{red}\fi}
\title{One-Forcing: Towards Stable One-Step\\Autoregressive Video Generation}
\author{%
  Jiaqi Feng\textsuperscript{1}\thanks{Equal contribution.} \quad
  Justin Cui\textsuperscript{2}\footnotemark[1] \quad
  Yuanhao Ban\textsuperscript{2} \quad
  Cho-Jui Hsieh\textsuperscript{2}\\
  \textsuperscript{1}Tsinghua University \quad
  \textsuperscript{2}UCLA\\[2.5ex]
  {\normalfont Project Page: \url{https://aurora-edu.github.io/one-forcing/}}\\[0.5ex]
  {\normalfont Code: \url{https://github.com/Aurora-edu/One-Forcing}}
}
\iclrfinalcopy
\hypersetup{pdftitle={One-Forcing: Towards Stable One-Step Autoregressive Video Generation},pdfauthor={Jiaqi Feng, Justin Cui, Yuanhao Ban, Cho-Jui Hsieh},hidelinks}

\makeatletter
\renewcommand{\@maketitle}{%
  \vbox{\hsize\textwidth
    \centering
    {\LARGE\scshape\@title\par}%
    \vskip 21pt
    \begin{tabular}[t]{c}\bfseries\rule{\z@}{24pt}\@author\end{tabular}\par
    \vskip 0.3in minus 0.1in
  }%
}
\makeatother

\begin{document}

\vspace*{10pt}
\maketitle
\lhead{Preprint}

\vspace{6pt}
\begin{abstract}
\vspace{3pt}
{
Recent advances in autoregressive diffusion-based video generation have substantially improved the quality of real-time video synthesis. However, most existing methods still require multiple denoising steps, while reducing sampling to a single step often leads to severe quality degradation: trajectory-based consistency distillation methods often produce videos with weak dynamics, whereas DMD-based methods, such as Self-Forcing, tend to generate blurry frames. We attribute this limitation to the teacher trajectories exhibiting highly concentrated curvature near the high-noise endpoint, which poses a fundamental geometric challenge for one-step distillation under the consistency distillation framework. To address these limitations, we propose One-Forcing, a simple yet effective approach that augments the DMD objective with an auxiliary GAN loss for high-quality and efficient one-step video generation. We further find that framewise autoregression stabilizes adversarial training, enabling higher-quality generation with substantially fewer training iterations than chunkwise autoregression. Experiments on VBench show that One-Forcing achieves a total score of 83.76, establishing state-of-the-art performance among one-step causal video generation methods while remaining competitive with strong multi-step approaches.
}
\end{abstract}

\vspace{16pt}
\begin{figure}[!ht]
\setlength{\abovecaptionskip}{16pt}
\centering
\includegraphics[width=\linewidth]{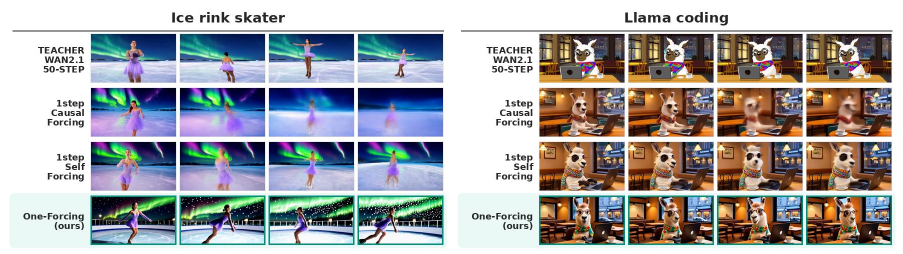}
\caption{Visualization results on two prompts: Wan2.1 teacher uses 50 denoising steps, Causal Forcing, Self Forcing, and \method{} use one-step autoregressive sampling. Our method exhibits excellent dynamics and visual quality.}
\label{fig:teaser}
\end{figure}

\clearpage
\section{Introduction}
Diffusion-based video generation has {achieved striking visual fidelity and complex spatiotemporal dynamics, as demonstrated by bidirectional models} such as Sora~\citep{openai2024sora}, Veo~\citep{deepmind_veo}, Wan~\citep{wan2025wan}, HunyuanVideo~\citep{kong2024hunyuanvideo}{,} and Seedance~\citep{seedance2026seedance}. {However, jointly denoising entire sequences incurs substantial computational costs and limits real-time interaction.} Autoregressive video generators address this limitation by producing frames or short temporal blocks {sequentially}~\citep{yin2025causvid,huang2025selfforcing,zhu2026causalforcing}, enabling {streaming} applications such as world simulation~\citep{ha2018worldmodels,hafner2024dreamerv3,parkerholder2024genie2,zhu2026astra} and interactive game engines~\citep{bruce2024genie,valevski2025gamengen}. {Yet} most causal video systems still require {multiple denoising steps} per block, {leaving the sampling budget as a key bottleneck} for end-to-end latency. The central question of this paper is {therefore} whether a causal video generator can preserve strong visual quality and motion dynamics {in the one-step regime}.

{
Existing fast distillation objectives leave a gap in this regime. {Consistency-style methods can produce high-quality videos with a few denoising steps, but often lose motion and structure when reduced to one step.} As shown in Figure~\ref{fig:curvature}, Wan~\citep{wan2025wan} video trajectories {exhibit sharply concentrated curvature near the high-noise endpoint: their strongest departure from a straight noise-to-data path occurs near the starting point of generation. This suggests a geometric challenge for predicting the final video directly from the highest-noise input, whereas multiple sampling steps allow subsequent predictions to refine the initial estimate.} Distribution Matching Distillation (DMD) offers {an alternative by matching output distributions without following a prescribed teacher trajectory}~\citep{yin2024dmd}. However, {in our one-step evaluations, Self Forcing still produces blurry outputs with missing fine detail. Recent analysis shows that low-frequency components can dominate DMD updates and delay the recovery of sharp boundaries and textures, offering a possible explanation for this behavior~\citep{zhou2026spectralamplitudepurificationdistribution}. Adversarial supervision can provide complementary feedback on fine details, as demonstrated at low noise levels in SDXL-Lightning~\citep{lin2024sdxllightningprogressiveadversarialdiffusion}. These observations motivate augmenting DMD with a discriminator trained against real videos to improve visual realism.} In autoregressive generation, {this supervision must remain effective as the model's own predictions become the context for subsequent frames.}}

To address these challenges, we propose One-Forcing, a framework for high-quality one-step autoregressive video generation. Building on the adversarial distribution matching formulation of DMD2~\citep{yin2024dmd2}, we complement teacher guidance with direct supervision from real videos. Making this supervision effective under self-generated context also requires stable adversarial learning. Here, we find that framewise autoregression plays a key role, enabling stable and fast adversarial learning while improving generation quality over its chunkwise counterpart. We hypothesize that finer temporal credit assignment helps adversarial feedback guide learning as the generated context evolves. This finding highlights the importance of considering adversarial supervision and autoregressive granularity together when distilling video generators to a single step. On VBench, One-Forcing achieves a total score of 83.76, outperforming the evaluated one-step causal video baselines and remaining competitive with strong many-step approaches. In summary, our contributions are:
\begin{itemize}
    \item We identify a geometric obstacle to one-step video distillation: video teacher trajectories exhibit sharply
  concentrated curvature near the high-noise endpoint, unlike image teachers commonly used in consistency distillation.
  This provides an explanation for why trajectory-based objectives degrade sharply when compressed to a single video
  generation step.
    \item {We propose \method{},  which combines DMD and adversarial learning for one-step autoregressive video generation. By reusing the fake-score network as a shared critic, our method introduces supervision from real videos without requiring a separate discriminator backbone.}
    \item {We demonstrate that framewise autoregression supports stable and fast adversarial learning and improves generation quality over its chunkwise counterpart, making it an important component of effective one-step video distillation.}
    \item On VBench, \method{} achieves \emph{state-of-the-art} one-step causal video generation (83.76 total) and remains competitive with strong many-step approaches.
\end{itemize}

\begin{figure}[t]
\centering
\includegraphics[width=\linewidth]{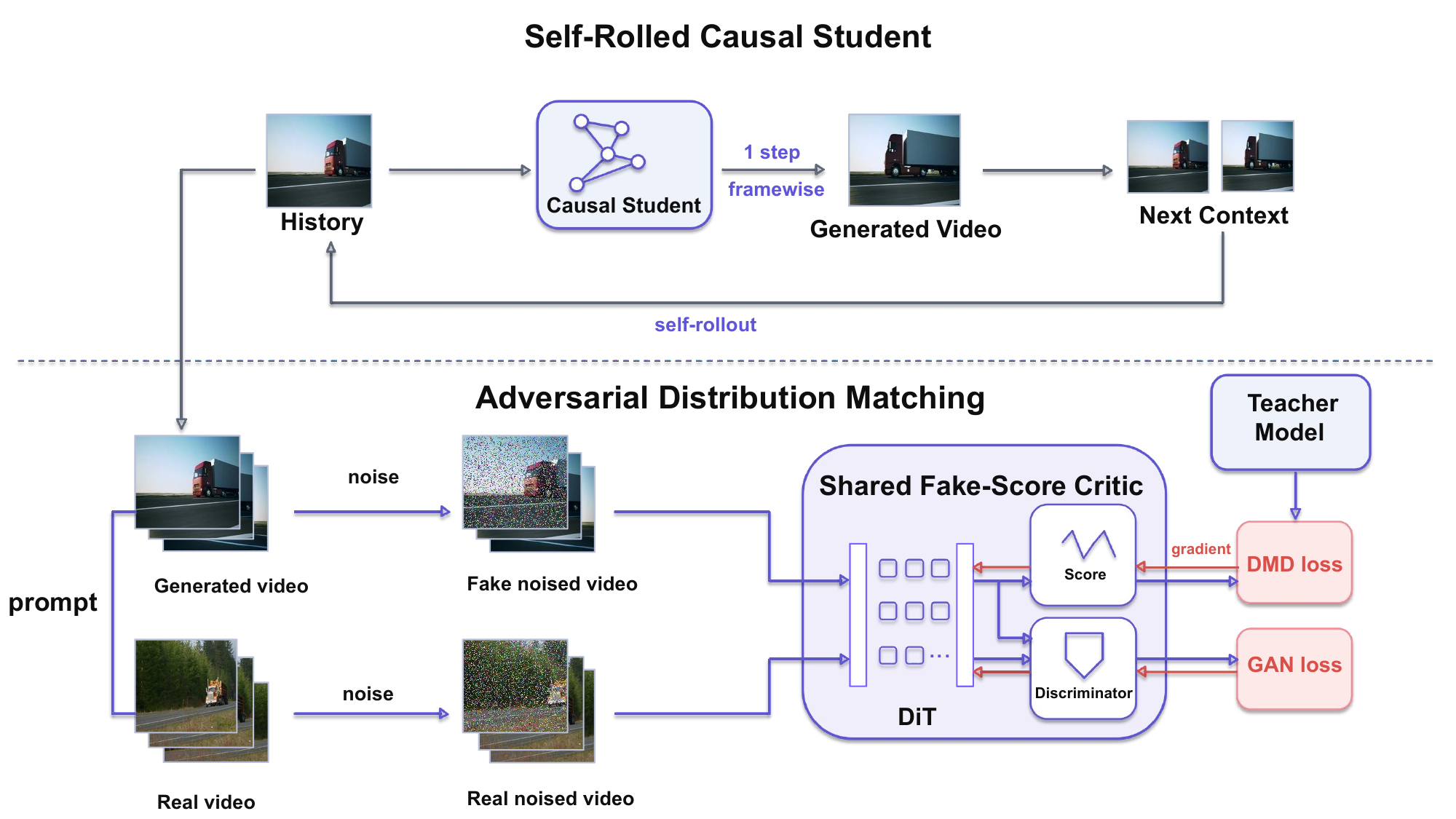}
\caption{\textbf{\method{} training framework.} {\method{} trains on} one-step {framewise autoregressive rollouts, combining teacher guidance through DMD and adversarial supervision from real videos with a shared} fake-score backbone.}
\label{fig:overview}
\end{figure}

\section{Related Works}

\subsection{Bidirectional and Autoregressive Video Generation}

Current video diffusion models fall into two paradigms. \emph{Bidirectional} models denoise an entire clip with full spatiotemporal attention, achieving strong coherence at {quadratic attention cost}~\citep{ho2022videodiffusion,ho2022imagenvideo,yang2024cogvideox,kong2024hunyuanvideo,wan2025wan}. {Full-clip processing limits their suitability for streaming.}

Autoregressive (causal) video generators {condition each new block on previously generated outputs through a KV cache, enabling real-time streaming.} Self Forcing~\citep{huang2025selfforcing} first demonstrated that training on self-generated context with a holistic video-level loss can close the train--test gap in causal video diffusion. Causal Forcing~\citep{zhu2026causalforcing} {uses an autoregressive teacher for ODE initialization to bridge the architectural gap between bidirectional and causal models}, yielding improvements in dynamics and instruction following. Causal Forcing++~\citep{zhao2026causalforcingplusplus} {replaces causal ODE initialization with causal consistency distillation, reducing preparation cost and enabling framewise two-step generation.} {These initialization strategies complement \method{}'s focus on one-step distribution matching.} Other notable systems include CausVid~\citep{yin2025causvid}, MAGI-1~\citep{teng2025magi}, LongLive~\citep{yang2025longlive}, Rolling Forcing~\citep{liu2026rollingforcing}, Infinity-RoPE~\citep{yesiltepe2025infiniterope}, and Self-Forcing++~\citep{cui2025selfforcingpp}. Despite these advances, {reducing sampling to one step still causes pronounced quality degradation.}

\subsection{Diffusion Distillation}

Two complementary approaches exist for compressing multi-step diffusion or flow models into fewer steps. One line relies on continuous-time transport trajectories between noise and data. Flow matching~\citep{lipman2022flow} and rectified flow transformers~\citep{esser2024scaling} learn velocity fields that parameterize such transport paths, while consistency distillation enforces that a student's prediction remains invariant along a teacher trajectory, typically a PF-ODE, enabling few-step or one-step generation~\citep{song2023consistency,song2023improved,lu2024scm}. Consistency-style and related trajectory-compression methods have scaled well for images~\citep{luo2023latent,frans2024one,zheng2025large} and been extended to video~\citep{Lv_2025_ICCV}, but they implicitly assume trajectories that are smooth enough to be faithfully compressed.

{Distribution Matching Distillation} (DMD)~\citep{yin2024dmd} {instead estimates a reverse-KL gradient from the difference between teacher and generated-distribution scores, matching distributions without following a specific teacher trajectory. DMD2~\citep{yin2024dmd2} further incorporates adversarial supervision from real data.} Recent video extensions {use windowed autoregressive rollouts~\citep{huang2025selfforcing,ge2026salt}}, reward-weighted distribution matching~\citep{lu2025rewardforcing}, and diagonal multi-step scheduling~\citep{liu2026diagonaldistill}.

\subsection{Adversarial Training for Video Generation}

Generative adversarial networks~\citep{goodfellow2014gan} offer single-pass generation by training a discriminator to separate real and generated samples. Early video GANs {used} 3D convolutions or {motion--appearance} decompositions~\citep{vondrick2016scenedynamics,tulyakov2018mocogan}, {while} later work improved temporal fidelity with continuous-time generators~\citep{skorokhodov2022styleganv}. {Recent methods use adversarial learning as a post-training or distillation signal for diffusion models.} Adversarial Diffusion Distillation (ADD)~\citep{sauer2024adversarial} {sharpens} one-step image outputs. Adversarial Post-Training (APT)~\citep{lin2025diffusion} {extends this approach} to one-step text-to-video generation. Autoregressive APT (AAPT)~\citep{lin2025autoregressive} combines adversarial training with student-forcing in a causal KV-cache architecture, generating {one} latent frame per forward pass {for real-time streaming}. Adversarial Self-Distillation~\citep{yang2026onestepcausalvideo} and Phased One-Step Adversarial Equilibrium~\citep{cheng2025pose} similarly {apply} adversarial objectives to few-step video generation. {Overall, these advances demonstrate that adversarial training can improve video quality while reducing generation to one or a few sampling steps.}

\section{Method}

\subsection{{Preliminaries}}

\paragraph{{Consistency distillation.}}
{Consistency distillation trains an endpoint predictor $f_\theta(x_t,t,c)$ to produce consistent clean predictions from different states along a teacher trajectory, conditioned on a text prompt $c$. Predictions at adjacent teacher states are encouraged to agree through}
\begingroup

\begin{equation}
    \mathcal{L}_{\mathrm{CD}}(\theta)
    = \mathbb{E}_{x_t,t,c}\left[
    \operatorname{MSE}\left(
    f_\theta(x_t,t,c),
    \stopgrad\left[
    f_{\bar{\theta}}\left(
    \Phi_{\Delta t}(x_t,t,c),t-\Delta t,c
    \right)\right]\right)\right],
    \label{eq:consistency-distillation}
\end{equation}
\endgroup
{where $\Phi_{\Delta t}$ advances the teacher trajectory from $t$ to $t-\Delta t$, $f_{\bar{\theta}}$ is an exponential moving average of the student, and $\stopgrad$ denotes stop-gradient. MSE denotes mean squared error. The resulting predictor supports direct noise-to-data generation and refinement through additional sampling steps~\citep{song2023consistency}.}

\paragraph{{Self Forcing.}}
{Self Forcing trains an autoregressive generator on its own rollouts, with each prediction conditioned on previously generated outputs through the KV cache~\citep{huang2025selfforcing}. Given a generated sequence $x_\theta=G_\theta(z,c)$, DMD supervises its distribution through a reverse-KL gradient~\citep{yin2024dmd}:}
\begingroup

\begin{equation}
    \nabla_{\theta}\mathrm{KL}(\pgen\Vert\pdata)
    =
    \mathbb{E}_{x_\theta}
    \left[
    J_{\theta}(z,c)^{\top}
    \left(
    \nabla_x \log \pgen(x_\theta\mid c)
    -
    \nabla_x \log \pdata(x_\theta\mid c)
    \right)
    \right],
    \label{eq:dmd-kl-gradient}
\end{equation}
\endgroup
{where $\pgen$ and $\pdata$ denote the generated and target distributions, respectively, and $J_\theta(z,c)$ is the generator Jacobian with respect to $\theta$. DMD estimates the score difference using a trainable fake-score model and a frozen teacher on noised rollouts. This objective matches output distributions without prescribing a teacher endpoint for each input noise sample.}

\subsection{{Challenges in One-Step Autoregressive Video Distillation}}

\paragraph{{Trajectory geometry challenges consistency distillation.}}
{Consistency distillation can produce high-quality videos with four denoising steps, but generation quality drops noticeably when sampling is reduced to one step. This behavior is also observed for rCM~\citep{nie2026tmd}. To investigate this gap, we examine the geometry of teacher denoising trajectories and its effect on distillation.}

{Following the trajectory analyses in TMD and rectified flow~\citep{nie2026tmd,liu2022flow}, we compare the local teacher velocity with the constant velocity connecting its noise and data endpoints. Specifically, we compute}
\begin{equation}
    C(t_i)=\frac{1}{d}
    \left\|
    \frac{x_{t_i}-x_{t_{i-1}}}{t_i-t_{i-1}}-(x_1-x_0)
    \right\|_2^2,
    \label{eq:trajectory-curvature}
\end{equation}
where $t=0$ and $t=1$ denote the data and highest-noise endpoints, respectively, and $d$ is the number of latent coordinates. {This quantity serves as a curvature proxy, measuring how much local trajectory velocity departs from linear evolution between the endpoints.}

\begin{figure}[t]
    \centering
    \includegraphics[width=0.6\linewidth]{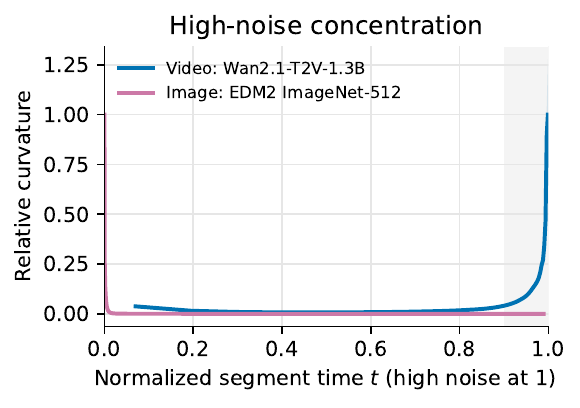}
    \caption{Relative trajectory-curvature profiles show high-noise concentration for Wan video generation but not for EDM2 ImageNet-512 image generation. Each curve is normalized by its own peak.}
    \label{fig:curvature}
\end{figure}

{Figure~\ref{fig:curvature} shows that Wan~\citep{wan2025wan} video trajectories concentrate $92.5\%$ of their measured curvature mass in the high-noise region $t\geq0.9$, whereas the EDM2 ImageNet-$512$~\citep{karras2024edm2} reference exhibits no comparable concentration. As shown in Appendix~\ref{app:curvature-details}, this pattern is consistent across prompts and persists when the measurement emphasizes temporal variation. The strongest departure from linear evolution therefore occurs near the starting point of video generation. This suggests a geometric challenge for one-step consistency distillation: four-step sampling allows later predictions to refine the initial estimate at lower noise levels, while one-step generation depends entirely on the endpoint prediction made from the highest-noise input.}

{To test whether trajectory geometry affects distillation quality, we preserve the teacher's noise and output endpoints and replace only its three intermediate states with time-aware linear interpolation. Table~\ref{tab:trajectory-rectification} shows that, with identical initialization, sample order, and optimization settings, rectification improves four-step VBench from $61.16$ to $71.11$ after $300$ training iterations. Because the endpoint pairs remain unchanged, this improvement shows that simplifying the intermediate trajectory helps the student learn the same endpoint mapping under a fixed training budget.}

\begin{table}[t]
\changedcolor
\centering
\small
\caption{VBench results for four-step consistency distillation with original and rectified teacher trajectories after 300 training steps.}
\label{tab:trajectory-rectification}
\begin{tabular}{l r r r}
\toprule
\textbf{Teacher trajectory} & \textbf{Total} & \textbf{Quality} & \textbf{Semantic} \\
\midrule
Original curved & 61.16 & 63.92 & 50.11 \\
Rectified & \textbf{71.11} & \textbf{75.37} & \textbf{54.05} \\
\bottomrule
\end{tabular}
\end{table}

{Together, the trajectory analysis and rectification experiment support a geometric explanation for the difficulty of compressing video consistency models to one step. They motivate direct output distribution matching, which allows the student to learn a noise-to-data mapping without reproducing the teacher's prescribed endpoint correspondence.}

\paragraph{{Fine-detail loss in one-step Self Forcing.}}
{Self Forcing provides this distributional alternative, but a substantial quality gap remains in the one-step setting. In our evaluation, its one-step outputs exhibit visible blur and loss of fine detail despite being trained on self-generated context, as shown in Figure~\ref{fig:teaser}. These imperfections also enter the conditioning context for subsequent predictions.}

{Recent analysis offers a possible explanation for this behavior. SAP-DMD finds that low spatial frequencies can dominate the DMD update direction, favoring coarse structure while delaying the learning of sharp boundaries and fine textures~\citep{zhou2026spectralamplitudepurificationdistribution}. SDXL-Lightning further reports that its adversarial loss emphasizes high-frequency details at low noise levels~\citep{lin2024sdxllightningprogressiveadversarialdiffusion}.}

{These observations motivate complementing teacher-derived distributional guidance with direct supervision from real videos. An adversarial objective provides an additional learning signal for visual realism as the student distribution evolves. \method{} combines this supervision with DMD and uses framewise autoregression to support stable learning from the resulting self-generated rollouts.}

\subsection{\method{}}

{Figure~\ref{fig:overview} summarizes \method{}. Starting from Self Forcing, we introduce real-video adversarial supervision alongside DMD and use framewise prediction to improve one-step generation. Both objectives supervise the videos produced under the student's own generated context.}

\paragraph{{Adversarial supervision with a shared backbone.}}
{We add a discriminator head $D_\phi$ to the trainable fake-score network, sharing its backbone for denoising and sequence-level real/fake discrimination. The head aggregates features from selected transformer layers into a scalar logit for the entire sequence. The bidirectional critic can evaluate both frame appearance and relationships across frames.}

{The discriminator compares real video latents with generated rollouts at matched noise levels. Real videos provide a direct reference for visual realism as the student distribution evolves. Denoting the noised real and generated sequences by $x_{\mathrm{real},t}$ and $x_{\theta,t}$, the adversarial losses are}
\begin{align}
    \mathcal{L}_{G}^{\mathrm{adv}}
    &=\mathbb{E}\left[\softplus\left(-D_\phi(x_{\theta,t},t,c)\right)\right],
    \label{eq:generator-adversarial-loss} \\
    \mathcal{L}_{D}^{\mathrm{adv}}
    &=\mathbb{E}\left[
    \softplus\left(-D_\phi(x_{\mathrm{real},t},t,c)\right)
    +\softplus\left(D_\phi(x_{\theta,t},t,c)\right)
    \right],
    \label{eq:discriminator-adversarial-loss}
\end{align}
{where $\softplus(u)=\log(1+\exp(u))$. Generated samples are detached when updating the discriminator.}

\paragraph{{Framewise autoregressive rollouts.}}
{To support stable learning with this joint supervision, we use framewise autoregressive rollouts. The generator predicts one latent frame at a time, conditioned on previously generated frames. Each prediction uses one denoising step and updates the KV cache before the next frame is generated. The student therefore learns with the evolving context produced by its own predictions.}

{Framewise autoregression decomposes generation into smaller conditional prediction problems and incorporates the newest generated content after every latent frame. Chunkwise generation jointly predicts several new frames before updating this context. This distinction becomes particularly relevant when each prediction has only one denoising step.}

{The adversarial objective evaluates the generated sequence jointly. We believe that this finer temporal decomposition helps assign sequence-level adversarial feedback to individual causal predictions, making learning more effective as the generated context evolves. Our experiments show that framewise generation achieves higher quality and converges in fewer training iterations than the chunkwise setting.}

\paragraph{{Joint objective and training.}}
{The generator combines DMD guidance with adversarial supervision:}
\begin{equation}
    \mathcal{L}_{G}
    =\mathcal{L}_{\mathrm{DMD}}
    +\lambda_G\mathcal{L}_{G}^{\mathrm{adv}}.
    \label{eq:joint-generator-loss}
\end{equation}
{The shared critic learns to estimate the generated-distribution score and discriminate real from generated videos:}
\begin{equation}
    {\mathcal{L}_{\mathrm{critic}}}
    =\mathcal{L}_{\mathrm{fake}}
    +\lambda_D\mathcal{L}_{D}^{\mathrm{adv}},
    \label{eq:joint-critic-loss}
\end{equation}
{where $\mathcal{L}_{\mathrm{fake}}$ denotes the fake-score denoising loss, and $\lambda_G$ and $\lambda_D$ control the adversarial contributions. DMD supplies teacher-derived distributional guidance, while the adversarial objective supplies direct supervision from real videos. Both act on framewise autoregressive rollouts.}

{We initialize the generator from the ODE-distilled causal checkpoint provided by Causal Forcing~\citep{zhu2026causalforcing}. Each training iteration updates the critic, and the generator is updated every five iterations using a separate minibatch. During generator updates, critic parameters are frozen while gradients propagate through its input. Generator gradients are restricted to the selected rollout window, and KV-cache updates are detached.}

\section{Experiments}

\subsection{Implementation Details}

Following~\citep{zhu2026causalforcing,huang2025selfforcing}, the generator is initialized from an ODE initialized checkpoint~\citep{zhu2026causalforcing} and uses one denoising timestep per autoregressive block. The trainable fake-score critic uses Wan2.1-T2V-1.3B, while the frozen real-score teacher uses Wan2.1-T2V-14B. Unless otherwise specified, the reported configuration uses 21 latent frames with 1 latent frame per autoregressive block for the framewise model and 3 latent frames per block for the chunkwise model, a generator update every five critic updates, and non-relativistic adversarial losses. We use 8$\times$H100 GPUs for training. Additional implementation details from the final training configuration are provided in Appendix~\ref{app:implementation-details}.

\subsection{Evaluation}

We {evaluate visual quality and text-video alignment using} VBench~\citep{huang2023vbench}, reporting the official normalized total, quality, and semantic scores. {Unless otherwise specified, our model generates videos at $832\times480$ resolution and 16 FPS, with one latent frame per autoregressive block.} {For inference, following ASD~\citep{yang2026onestepcausalvideo}, we adopt First-Frame Enhancement (FFE) to improve the quality of the first frame.} {Inference requires a single H100 GPU.} {Additional evaluation settings and prompt-processing details are provided in Appendix~\ref{app:implementation-details}.}

\paragraph{{Baselines.}}
{Our primary comparisons are with one-step autoregressive video generators, including} Self Forcing~\citep{huang2025selfforcing}, Causal-Forcing~\citep{zhu2026causalforcing}, and ASD~\citep{yang2026onestepcausalvideo}. {These methods use the same Wan2.1-1.3B generator backbone as \method{}, allowing comparison at a common model scale.} {To assess how closely one-step generation approaches multi-step quality, we also include} Wan2.1~\citep{wan2025wan}, LTX-Video~\citep{hacohen2024ltxvideo}, SkyReels-V2~\citep{chen2025skyreelsv2}, NOVA~\citep{deng2025nova}, Pyramid Flow~\citep{jin2025pyramidflow}, MAGI-1~\citep{teng2025magi}, CausVid~\citep{yin2025causvid}, and {four-step} Self Forcing. {Model sizes, output resolutions, and sampling steps are summarized in Table~\ref{tab:vbench-main}.}

\subsection{Main Results}

\begin{table}[t]
\centering
\small
\setlength{\tabcolsep}{4.2pt}
\renewcommand{\arraystretch}{1.08}
\caption{VBench results for one-step and many-step video generation. }
\label{tab:vbench-main}
\begin{adjustbox}{max width=0.98\linewidth}
\begin{tabular}{l c c c c c c}
\toprule
\textbf{Model} & \textbf{\#Params} & \textbf{Resolution} & \textbf{NFE}
& \multicolumn{3}{c}{\textbf{VBench Scores} $\uparrow$} \\
\cmidrule(l){5-7}
& & &
& \makecell[c]{\textbf{Total}}
& \makecell[c]{\textbf{Quality}}
& \makecell[c]{\textbf{Semantic}} \\
\midrule

\rowcolor{gray!12}
\multicolumn{7}{l}{\textit{Multiple steps}} \\
MAGI-1~\citep{teng2025magi}
    & 4.5B & $832\times480$ & 64 & 79.18 & 82.04 & 67.74 \\
Wan2.1~\citep{wan2025wan}
    & 1.3B & $832\times480$ & 50 & \textbf{84.26} & \textbf{85.30} & \textbf{80.09} \\
SkyReels-V2~\citep{chen2025skyreelsv2}
    & 1.3B & $960\times540$ & 30 & 82.67 & 84.70 & 74.53 \\
NOVA~\citep{deng2025nova}
    & 0.6B & $768\times480$ & 25 & 80.12 & 80.39 & 79.05 \\
LTX-Video~\citep{hacohen2024ltxvideo}
    & 1.9B & $768\times512$ & 20 & 80.00 & 82.30 & 70.79 \\
Pyramid Flow~\citep{jin2025pyramidflow}
    & 2B & $640\times384$ & 20 & 81.72 & 84.74 & 69.62 \\
CausVid~\citep{yin2025causvid}
    & 1.3B & $832\times480$ & 4 & 81.18& 84.41&68.30 \\
Self Forcing~\citep{huang2025selfforcing}
    & 1.3B & $832\times480$ & 4 &  83.46& 84.77 & 78.24 \\
\midrule

\rowcolor{gray!12}
\multicolumn{7}{l}{\textit{Single step}} \\
Self Forcing~\citep{huang2025selfforcing}
    & 1.3B & $832\times480$ & 1 & 77.18&  79.40&  68.34  \\
Causal-Forcing~\citep{zhu2026causalforcing} & 1.3B & $832\times480$ & 1 & 78.39& 80.67& 69.25 \\
ASD~\citep{yang2026onestepcausalvideo}
    & 1.3B & $832\times480$ & $1$ & 79.12& 81.35& 70.19 \\
Ours(chunkwise)& 1.3B & $832\times480$ & $1$ & 81.60& 83.65& 73.41 \\
Ours(framewise)
    & 1.3B & $832\times480$ & $1$ & \textbf{83.76} & \textbf{85.22}& \textbf{77.91} \\
\bottomrule
\end{tabular}
\end{adjustbox}
\end{table}

Table~\ref{tab:vbench-main} summarizes the results. In the one-step setting, \method{} (framewise) achieves a total score of 83.76 with quality 85.22 and semantic 77.91, outperforming all prior one-step causal methods including Self Forcing, Causal-Forcing, and ASD by 4--7 points in total score. Notably, with only a single {denoising step}, \method{} \emph{surpasses most many-step baselines} that use 4 to 25 denoising steps, including Self Forcing, CausVid, LTX-Video, Pyramid Flow, and NOVA. The remaining gap to the {50-step Wan2.1-1.3B baseline} is marginal, showing that one-step generation can approach multi-step quality given an effective distillation objective. {The framewise model, which generates one latent frame per autoregressive block, outperforms the chunkwise variant with three-frame blocks, achieving a VBench total score of 83.76 versus 81.60. Beyond improving generation quality, framewise autoregression also simplifies adversarial training. In our experiments, the framewise configuration trains stably without R1/R2 regularization, whereas the chunkwise counterpart requires these penalties to maintain stability. This avoids the additional computational overhead associated with discriminator regularization. Under their respective stable training configurations, the framewise model reaches its reported performance after 200 distillation iterations, compared with 750 iterations for the chunkwise model.} Our chunkwise model (81.60) nonetheless surpasses all existing one-step methods, confirming that the proposed objective is effective across different block {sizes}.

\paragraph{\changedtext{Long video generation.}}
{ We compare one-step models on 20-second videos to assess longer autoregressive rollouts. As shown in Table~\ref{tab:long-video}, \method{} outperforms Self Forcing on VBench, with improvements in quality and semantic alignment. These results extend its benefits beyond short-clip generation.
}

\begin{table}[h]
\changedcolor
\centering
\small
\caption{VBench results for 20-second video generation. Both models use one-step inference. }
\label{tab:long-video}
\begin{tabular}{l r r r r}
\toprule
\textbf{Model} & \textbf{Total} & \textbf{Quality} & \textbf{Semantic} & \textbf{Multiple objects} \\
\midrule
Self Forcing & 75.30 & 77.47 & 66.61 & 39.33 \\
\method{} & \textbf{78.52} & \textbf{79.38} & \textbf{75.11} & \textbf{67.91} \\
\bottomrule
\end{tabular}
\end{table}

\subsection{Ablation}
{In this section, we study the effects of adversarial supervision, FFE, and initialization on generation quality, and examine discriminator behavior during training.}
\begin{table}[h]
\changedcolor
\centering
\small
\caption{VBench ablations of adversarial training and FFE. More details are in Appendix~\ref{app:implementation-details}.}
\label{tab:gan-ffe-ablation}
\setlength{\tabcolsep}{3.5pt}
\begin{minipage}[t]{0.48\linewidth}
\centering
\parbox[c][\baselineskip][c]{\linewidth}{\centering\textit{(a) Adversarial objective}}\par\smallskip
\begin{tabular}{@{}l r r r@{}}
\toprule
\textbf{Configuration} & \textbf{Total} & \textbf{Quality} & \textbf{Semantic} \\
\midrule
DMD-only & 75.27 & 78.29 & 63.19 \\
DMD+GAN{(Ours)} & \textbf{80.36} & \textbf{83.22} & \textbf{68.92} \\
\bottomrule
\end{tabular}
\end{minipage}\hfill
\begin{minipage}[t]{0.48\linewidth}
\centering
\parbox[c][\baselineskip][c]{\linewidth}{\centering\textit{(b) First-Frame Enhancement (FFE)}}\par\smallskip
\begin{tabular}{@{}l r r r@{}}
\toprule
\textbf{Configuration} & \textbf{Total} & \textbf{Quality} & \textbf{Semantic} \\
\midrule
Without FFE & 80.69 & 81.15 & \textbf{78.84} \\
With FFE {(Ours)} & \textbf{83.30} & \textbf{84.65} & 77.93 \\
\bottomrule
\end{tabular}
\end{minipage}
\end{table}

\paragraph{\changedtext{Adversarial objective and FFE.}}
\changedtext{{As shown in Table~\ref{tab:gan-ffe-ablation}a, removing} the adversarial branch while retaining DMD and FFE degrades generation quality. This supports our central design of complementing score matching with real-data discrimination, and shows that the gain from adversarial training is not attributable to FFE. The FFE ablation {in Table~\ref{tab:gan-ffe-ablation}b shows that our method outperforms the evaluated prior one-step baselines even without FFE. Consistent with ASD~\citep{yang2026onestepcausalvideo}, enhancing the first block, which conditions subsequent generation, provides an inference-time complement to our training objective.}}

\begin{table}[t]
\centering
\small
\setlength{\tabcolsep}{5pt}
\renewcommand{\arraystretch}{1.08}
\caption{{Ablation study of Causal ODE Initialization and Consistency Distillation Initialization on VBench.} }
\label{tab:ablation}
\begin{tabular}{l c c c}
\toprule
\textbf{Configuration} & \textbf{Total} & \textbf{Quality} & \textbf{Semantic} \\
\midrule
\textbf{Causal ODE Initialization(Ours)} & \textbf{83.76} & \textbf{85.22} & \textbf{77.91} \\
Consistency Distillation Initialization & 82.36 & 82.82 & 80.50\\
\bottomrule
\end{tabular}
\end{table}

\paragraph{Initialization strategy.}
We compare two framewise configurations that differ in initialization: causal ODE initialization (row 1) and causal Consistency Distillation initialization (row 2). Both use 1-frame blocks. The ODE-initialized model achieves a higher total score (83.76 vs.\ 82.36) and substantially stronger dynamic degree (52.76 vs.\ 23.61), while the CD-initialized variant obtains better semantic scores (80.50 vs.\ 77.91). We attribute the dynamic advantage of ODE initialization to its training data containing richer motion information from the multi-step ODE trajectory, which provides the generator with a stronger motion prior during distillation. The CD initialization, on the other hand, starts from a model already trained for consistency, which benefits semantic alignment but tends to suppress large motions.

\begin{wrapfigure}{r}{0.42\textwidth}
\centering
\includegraphics[width=\linewidth]{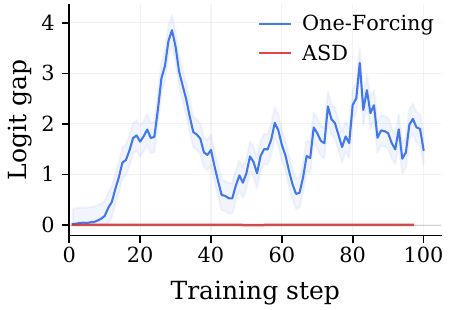}
\caption{Discriminator logit gap {between reference and generated videos} during training. \method{} {maintains a clear gap}, while ASD stays near zero.}
\label{fig:gan-stability}
\end{wrapfigure}

\paragraph{Discriminator effectiveness.}
We compare the adversarial training dynamics of \method{} against ASD~\citep{yang2026onestepcausalvideo}. Both methods attach a classification branch to the fake-score backbone operating in noised latent space, but they differ fundamentally in what constitutes the ``real'' distribution for the discriminator. \method{} trains the discriminator on \emph{real data}: noised samples from actual videos in the training set, providing a fixed, high-quality target distribution. ASD instead uses a \emph{self-distillation} target, where the output of an $(n{+}1)$-step model serves as ``real'' for the $n$-step student, meaning the discriminator must distinguish between two imperfect model outputs rather than between generated and genuine data.

Figure~\ref{fig:gan-stability} plots the discriminator logit gap $|l_r - l_f|$ {during training, where $l_r$ and $l_f$ denote the discriminator logits for reference and generated videos, respectively.} \method{} maintains a large, actively varying gap ($\mu{=}1.53$, $\sigma{=}1.20$): the distributional distance between generated latents and real data is substantial and evolves as the generator improves, indicating a healthy adversarial dynamic. In contrast, ASD's logit gap stays near zero ($\mu{=}0.001$, $\max{<}0.006$) from the very first steps. Because both sides of ASD's comparison are model-generated latents with minimal distributional difference, the discriminator never receives a meaningful learning signal and effectively collapses. This confirms that grounding the adversarial signal in \emph{real data} is critical for effective GAN-based video distillation.

\section{Conclusion}

{We presented \method{}, a framework for high-quality one-step autoregressive video generation that combines DMD with adversarial supervision from real videos. Our trajectory analysis reveals concentrated curvature near the high-noise endpoint, providing a geometric perspective on the difficulty of one-step consistency distillation. Within this framework, we identify framewise autoregression as an important ingredient for stable and fast adversarial learning, achieving higher generation quality than its chunkwise counterpart. On VBench, \method{} achieves a total score of 83.76, outperforming the evaluated one-step causal video baselines and remaining competitive with strong many-step approaches.}

\section{\changedtext{Limitations} {and Future Work}}

\method{} requires real data as the ``real'' distribution for the discriminator, which is different from data-free methods such as Self Forcing~\citep{huang2025selfforcing} and ASD~\citep{yang2026onestepcausalvideo}. However, these data are already available in standard forcing-like distillation settings where training videos or their precomputed representations are used. {Future work will explore larger backbones, efficient attention, and long-video generation frameworks~\citep{yang2025longlive,liu2026rollingforcing,yesiltepe2025infiniterope,cui2025selfforcingpp} for higher-resolution and longer-duration generation. Other directions include adaptive step scheduling that allocates more denoising steps to perceptually complex segments, and action-conditioned generation for interactive world modeling~\citep{deepmind2025genie3,bruce2024genie,sun2025worldplay,zhang2025matrixgame}.}

\subsection*{\changedtext{AI use statement}}
\changedtext{We used generative AI tools to assist with literature review, manuscript drafting and editing, refinement of research hypotheses, interpretation of experimental results, code implementation, \LaTeX{} formatting, and figure preparation. We also used Qwen2.5-7B-Instruct to rewrite evaluation prompts, as described in Appendix~\ref{app:implementation-details}. The authors take responsibility for the final content of this work.}

\subsection*{\changedtext{Ethics Statement}}
\changedtext{Efficient autoregressive video generation can support creative content production, accessibility tools, rapid prototyping, simulation, and interactive world modeling. However, it also lowers barriers to creating fabricated media, spreading disinformation, impersonating individuals, and reproducing harmful stereotypes or biases in training data. Responsible deployment should therefore include dataset and prompt filtering, provenance tracking, watermarking or content credentials, synthetic-media detection, clear disclosure of generated content, and usage restrictions in sensitive domains. These safeguards require dedicated evaluation alongside improvements in generation quality and inference speed.}

\subsection*{\changedtext{Reproducibility statement}}
\changedtext{Appendix~\ref{app:implementation-details} describes the model initialization, noise schedule, training objectives, hyperparameters, and inference procedure. Appendix~\ref{app:curvature-details} provides the trajectory-curvature sampling and robustness details, and Appendices~\ref{app:loss-curves} and~\ref{app:vbench-radar} report training curves and detailed evaluation results.}

\bibliographystyle{iclr2027_conference}
\bibliography{references}

\clearpage
\appendix
\section{Details of Implementations}
\label{app:implementation-details}

Our implementation is based on the Causal Forcing codebase~\citep{zhu2026causalforcing} and the Wan2.1 model family~\citep{wan2025wan}. The reported framewise \method{} model is initialized from an ODE-trained causal model. The real-score network is a frozen bidirectional Wan2.1-T2V-14B model, and the trainable fake-score network is initialized from Wan2.1-T2V-1.3B. We reuse the fake-score backbone as the adversarial discriminator by adding register tokens, lightweight attention blocks, and a classification head to selected transformer layers. No {separate pixel-space} discriminator is used.

\paragraph{Noise schedule and model parameterization.}
Following Wan2.1, we use a flow-matching scheduler. For a sampled timestep $t\in[0,1000]$, the shifted noise level is
\[
    \sigma_t =
    \frac{k(t/1000)}{1+(k-1)(t/1000)},
\]
with shift factor $k=5$ for generator rollouts and for the DMD/GAN critic timestep sampling. The forward corruption process is
\[
    x_t=(1-\sigma_t)x_0+\sigma_t\epsilon,\qquad
    \epsilon\sim\mathcal{N}(0,I),
\]
and the flow-prediction target is $\epsilon-x_0$. During generation, the model predicts $v_\theta(x_t,t,c)$ and converts it to a clean latent estimate by $\hat{x}_0=x_t-\sigma_t v_\theta(x_t,t,c)$. Training rollouts use a single denoising timestep per autoregressive block.

\paragraph{Data and prompt processing.}
The distillation stage uses precomputed training data. Each training example contains a text prompt and the corresponding real data sample; raw videos are not decoded or reloaded during this stage. The adversarial real samples are drawn from this dataset, while fake samples are produced by the current one-step causal generator. The reported framewise model is trained on 21 latent frames with spatial latent size $60\times104$ and 16 latent channels, corresponding to $832\times480$ video generation. Text embeddings are computed with the frozen Wan text encoder, and classifier-free guidance uses the standard Wan negative prompt. \changedtext{For VBench evaluation, we rewrite the test prompts using Qwen/Qwen2.5-7B-Instruct following previous works~\citep{huang2025selfforcing,cui2025selfforcingpp}, except for the {trajectory-rectification experiment in Table~\ref{tab:trajectory-rectification}, the GAN-objective experiment in Table~\ref{tab:gan-ffe-ablation}a,} {and the 20-second video evaluation}, which use original prompts. Rewriting is held fixed within each paired comparison, including both FFE rows in Table~\ref{tab:gan-ffe-ablation}b.}

\paragraph{\changedtext{GAN and FFE evaluation.}}
\changedtext{The DMD-only and DMD+GAN comparison in Table~\ref{tab:gan-ffe-ablation}a uses step-200 raw generator weights without EMA, matched initialization, framewise training, and the same FFE inference schedule. Both are evaluated on the same 944 original prompts and generation seeds, with one generated sample per prompt and all 16 VBench dimensions using the normalized scoring formula. These scores use a one-sample protocol rather than the five-sample-per-prompt leaderboard protocol. The FFE comparison in Table~\ref{tab:gan-ffe-ablation}b keeps the same step-200 checkpoint and prompt set, changing only the inference schedule.}

\paragraph{Training details.}
We train the fake-score critic with the flow denoising objective on generated latents and train the adversarial branch to distinguish noised real data from noised generated samples. Each training iteration performs one critic update; every five iterations, we additionally update the generator on a separately sampled minibatch using the DMD surrogate and the adversarial generator loss. For DMD, the real-score model is evaluated with classifier-free guidance scale $5.0$, while the fake-score model is evaluated without classifier-free guidance. The DMD gradient is normalized by the average absolute real-score residual. We use AdamW for both generator and critic, mixed precision, gradient checkpointing, and full-shard FSDP. The reported framewise training run uses 8 NVIDIA H100 GPUs with a per-GPU batch size of 1.

\begin{algorithm}[htb]
\caption{\method{} Training}
\label{alg:one-forcing}
\begin{algorithmic}[1]
\REQUIRE Generator $G_\theta$, fake-score network ${v_\phi}$ with discriminator head $D_\phi$, frozen {teacher $v_T$}, Dataset $\mathcal{D}$, generator interval $K$, weights $\lambda_G$, $\lambda_D$
\FOR{training iteration $i=0,1,\ldots$}
    \IF{$i \bmod K = 0$}
        \STATE Sample prompt $c_G$ from $\mathcal{D}$
        \STATE Generate fake samples $x_\theta \leftarrow G_\theta(\epsilon_G, c_G)$ with one-step {framewise autoregressive} rollout
        \STATE Compute $\mathcal{L}_{\mathrm{DMD}}$ {from teacher and fake-score predictions on noised rollouts}
        \STATE Compute $\mathcal{L}_G^{\mathrm{adv}}$ with $D_\phi$ on independently noised fake samples
        \STATE Update $\theta$: $\mathcal{L}_G = \mathcal{L}_{\mathrm{DMD}} + \lambda_G \mathcal{L}_G^{\mathrm{adv}}$
    \ENDIF
    \STATE Sample a critic minibatch with prompt $c_\phi$ and real data $x_{\mathrm{real}}$ from $\mathcal{D}$
    \STATE Generate fake samples $\tilde{x}_\theta \leftarrow G_\theta(\tilde{\epsilon}, c_\phi)$ without generator gradients
    \STATE Train ${v_\phi}$ to denoise noised generated samples, giving $\mathcal{L}_{\mathrm{fake}}$
    \STATE Train $D_\phi$ to classify noised real data as real and noised generated samples as fake, giving $\mathcal{L}_D^{\mathrm{adv}}$
    \STATE Update $\phi$: ${\mathcal{L}_{\mathrm{critic}}} = \mathcal{L}_{\mathrm{fake}} + \lambda_D \mathcal{L}_D^{\mathrm{adv}}$
\ENDFOR
\end{algorithmic}
\end{algorithm}

\paragraph{Inference details.}
The reported framewise model generates one latent frame per autoregressive block. For the one-step setting, subsequent autoregressive blocks use one denoising update. Following \citep{yang2026onestepcausalvideo}, the first block is generated with a short four-step warm-up schedule to initialize the KV cache before one-step streaming continues. Unless otherwise specified, videos are decoded at $832\times480$ resolution and 16 FPS.

\begin{table}[htb]
\centering
\small
\setlength{\tabcolsep}{5pt}
\renewcommand{\arraystretch}{1.12}
\caption{Training hyperparameters for the reported framewise \method{} configuration.}
\label{tab:implementation-hyperparameters}
\begin{adjustbox}{max width=\textwidth}
\begin{tabular}{l l}
\toprule
\textbf{Hyperparameter} & \textbf{Value} \\
\midrule
Generator initialization & ODE-trained Causal model \\
Generator / real score / fake score & Wan2.1-T2V-1.3B / Wan2.1-T2V-14B / Wan2.1-T2V-1.3B \\
Objective & Flow denoising for fake score; DMD + noised-latent GAN for generator \\
Training frames & 21 latent frames, $16\times60\times104$ per frame \\
Frames per autoregressive block & 1 \\
Training rollout steps per block & 1 \\
Guidance scale & $5.0$ for {real-score guidance; no CFG for generator or fake score} \\
Timestep range and shift & 1000 training timesteps; shift factor $5.0$ for sampled DMD/GAN timesteps \\
Update schedule & One critic update per iteration; one generator update every five iterations \\
Optimizer & AdamW, $\beta_1=0$, $\beta_2=0.999$, weight decay $0.01$ \\
Learning rates & $1.0{\times}10^{-5}$ for generator and fake-score critic \\
Batch size & 1 per GPU on 8 GPUs \\
EMA & Decay $0.99$, starting after 50 iterations \\
GAN branch & Layers $\{21,29\}$, 2 registers, 1536 feature dim, 2048 FFN dim, 12 heads \\
GAN loss & Non-relativistic logistic loss, $\lambda_G=\lambda_D=0.03$ \\
Discriminator regularization & None \\
Systems & Mixed precision, gradient checkpointing, full-shard FSDP, activation CPU offload \\
Convergence steps & 200 iterations \\
\bottomrule
\end{tabular}
\end{adjustbox}
\end{table}

\section{Trajectory Curvature Analysis Details}
\label{app:curvature-details}

This appendix provides the sampling and robustness details for the trajectory-curvature calculation in Equation~\ref{eq:trajectory-curvature}. The video comparison uses 100 50-step Wan2.1-T2V-1.3B~\citep{wan2025wan} trajectories with diverse motion and scene prompts, shift $8$, classifier-free guidance $6$, and one deterministic seed per prompt. The image-domain comparison uses eight 256-step trajectories from the official EDM2~\citep{karras2024edm2} ImageNet-512 teacher, matching the teacher family used by scalable consistency models~\citep{lu2024scm}. EDM2 noise levels are normalized so $t=1$ is the highest-noise endpoint.

The high-noise concentration is stable across prompts: per-prompt estimates give $92.49\%\pm0.13\%$ curvature mass at $t\ge0.9$ (mean $\pm$ SEM; 95\% bootstrap CI $[92.24,92.73]\%$) and a high-noise/mid-noise ratio of $33.1\pm0.7$ (95\% bootstrap CI $[31.8,34.4]$). A temporal-difference version of the same metric, which removes static appearance and emphasizes motion structure, still places $88.6\%$ of the curvature mass at $t\ge0.9$ with a high/mid ratio of $19.3$.

\changedtext{The additional GAN-objective and trajectory-rectification training runs use 8$\times$H200 GPUs and a single training seed. Their paired evaluations use one generated sample per prompt; they do not estimate variation across training seeds.}

\paragraph{\changedtext{Four-step rectification evaluation.}}
\changedtext{The experiment in Table~\ref{tab:trajectory-rectification} uses two 300-step checkpoints trained on the same teacher-trajectory dataset. Only the three intermediate states differ; endpoints, initialization, sample order, and optimization are fixed. Both checkpoints are evaluated with four-step sampling, using raw generator weights without EMA on the same 944 original VBench prompts and generation seeds, with one sample per prompt and all 16 dimensions. This uses the normalized VBench scoring formula, but is not the five-sample-per-prompt leaderboard protocol. Rectification improves total score from 61.16 to 71.11, quality from 63.92 to 75.37, and semantic score from 50.11 to 54.05.}

\section{Training Loss Curves}
\label{app:loss-curves}

\begin{figure}[htb]
\centering
\includegraphics[width=\textwidth]{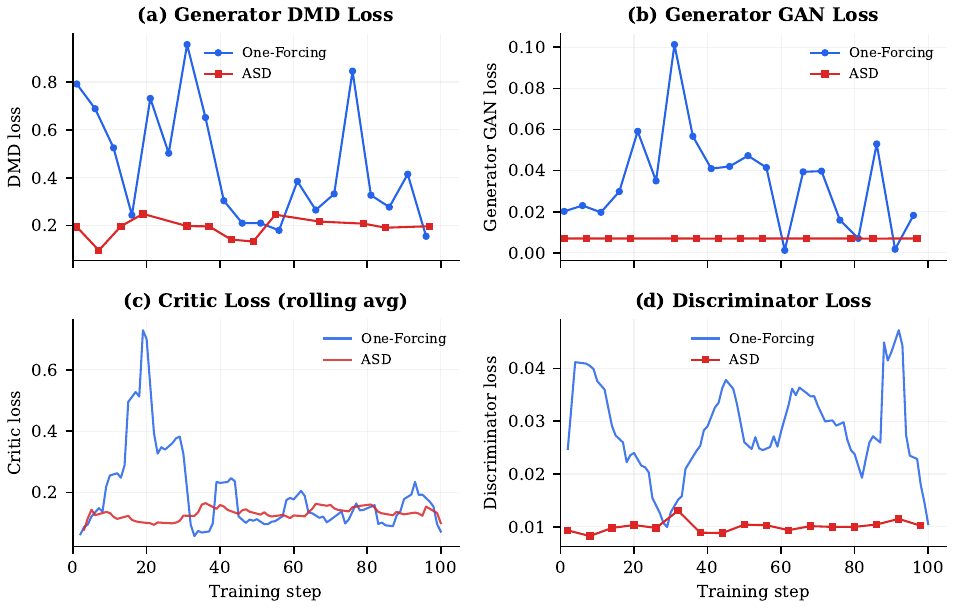}
\caption{Training loss curves for \method{} (blue) and ASD (red) over the first 100 steps. (a)~DMD loss. (b)~Generator GAN loss. (c)~Critic loss (rolling average). (d)~Discriminator loss.}
\label{fig:loss-curves}
\end{figure}

Figure~\ref{fig:loss-curves} compares the training loss curves of our \method{} and ASD~\citep{yang2026onestepcausalvideo} over the first 100 steps. Both methods start from an ODE-initialized generator checkpoint. Panel~(a) shows the DMD loss, which drives the score-matching component; both methods exhibit similar initial magnitudes, though \method{} stabilizes at a lower level. Panel~(b) reveals the generator GAN loss: \method{}'s loss varies actively as the discriminator provides meaningful gradients, whereas ASD's GAN loss flatlines at $\ln 2 \cdot 0.01 \approx 0.0069$ throughout training. Panels~(c) and~(d) show the critic and discriminator losses, respectively; \method{}'s discriminator loss decreases over training as it learns to distinguish real from fake, while ASD's discriminator loss remains constant.

\section{\changedtext{Training Beyond 200 Steps}}
\label{app:extended-training}

\changedtext{{To assess stability under extended training, we compare the main 200-step framewise \method{} result from Table~\ref{tab:vbench-main} with an additional evaluation at step 400. This evaluation uses Qwen-rewritten conditioning prompts, with five samples per prompt and all 16 VBench dimensions evaluated.} {As shown in Table~\ref{tab:extended-training}, the total score changes only slightly from 83.76 at step 200 to 83.93 at step 400, while the quality and semantic scores remain comparable. These results show that \method{} maintains stable generation quality even when we double the training budget.}}

\begin{table}[ht]
\changedcolor
\centering
\small
\caption{VBench scores during extended framewise training.}
\label{tab:extended-training}
\begin{tabular}{r r r r}
\toprule
\textbf{Training step} & \textbf{Total} & \textbf{Quality} & \textbf{Semantic} \\
\midrule
200 & {83.76} & {85.22} & {77.91} \\
400 & {83.93} & {85.41} & {78.04} \\
\bottomrule
\end{tabular}
\end{table}

\section{VBench Scores Across All Dimensions}
\label{app:vbench-radar}

\begin{figure}[htb]
\centering
\includegraphics[width=0.72\textwidth]{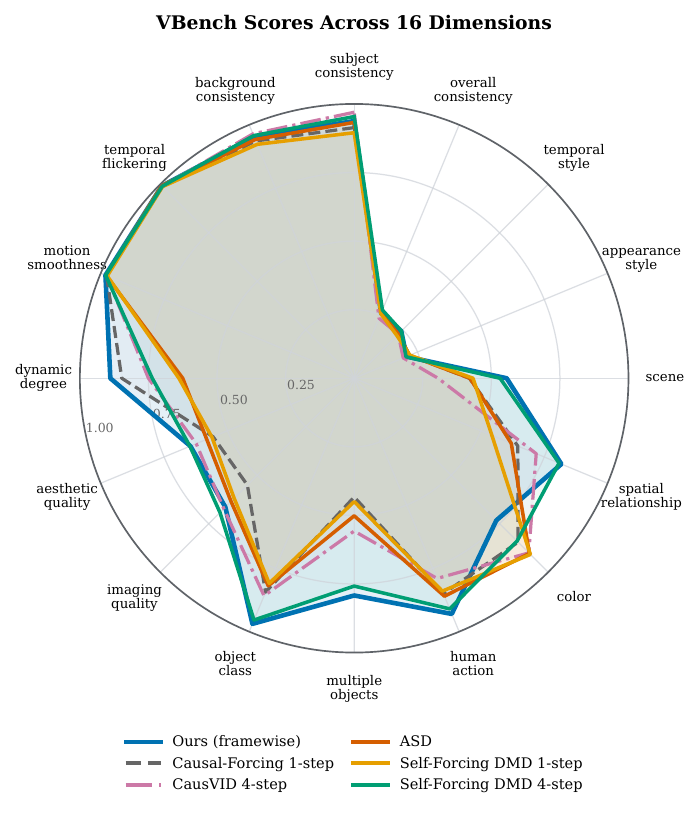}
\caption{VBench scores across all 16 dimensions for selected Table~\ref{tab:vbench-main} entries. Higher radial values indicate better normalized VBench sub-metric scores.}
\label{fig:vbench-radar}
\end{figure}

Following the evaluation visualization style of Self Forcing, Figure~\ref{fig:vbench-radar} expands the full 16-dimensional VBench profile for the Table~\ref{tab:vbench-main} entries with available per-dimension records: \method{} (framewise), Causal-Forcing 1-step~\citep{zhu2026causalforcing}, CausVID 4-step~\citep{yin2025causvid}, ASD~\citep{yang2026onestepcausalvideo}, Self Forcing DMD 1-step~\citep{huang2025selfforcing}, and Self Forcing DMD 4-step. \method{} improves over the one-step causal baselines in the aggregate score and shows stronger object, spatial-relation, scene, and dynamic-degree performance than ASD, Causal-Forcing, and one-step Self Forcing, while keeping high temporal smoothness. Compared with four-step Self Forcing, \method{} has a higher normalized VBench total and quality score, with gains in dynamic degree and several object/action/scene dimensions, while remaining slightly lower on color, imaging quality, and some consistency-style dimensions.

\section{\changedtext{Additional VBench Ablation Dimensions}}
\label{app:additional-ablation-dimensions}

\changedtext{Table~\ref{tab:ablation-extra-dimensions} expands the trajectory-rectification, adversarial-objective, and FFE comparisons in Tables~\ref{tab:trajectory-rectification} and~\ref{tab:gan-ffe-ablation} with additional VBench dimensions. Trajectory rectification improves subject consistency, motion smoothness, and imaging quality, connecting the geometric intervention to both temporal and visual quality. The adversarial objective improves subject consistency and imaging quality alongside multiple-object composition, while FFE improves subject consistency and multiple-object composition. These measurements clarify which aspects of generation contribute to the aggregate gains. They use the same paired settings as their corresponding main-text ablations: original prompts for rectification and the GAN comparison, and rewritten prompts for the FFE comparison.}

\begin{table}[ht]
\changedcolor
\centering
\small
\caption{Additional VBench dimensions for the main-text ablations. The baseline/variant pairs are original/rectified trajectories, DMD-only/DMD+GAN, and without/with FFE, respectively.}
\label{tab:ablation-extra-dimensions}
\begin{tabular}{l l r r}
\toprule
\textbf{Ablation} & \textbf{Metric} & \textbf{Baseline} & \textbf{Variant} \\
\midrule
Trajectory rectification & Subject consistency & 71.18 & \textbf{81.37} \\
 & Motion smoothness & 86.00 & \textbf{94.09} \\
 & Imaging quality & 47.78 & \textbf{53.80} \\
\midrule
Adversarial objective & Multiple objects & 39.41 & \textbf{65.24} \\
 & Subject consistency & 81.54 & \textbf{92.93} \\
 & Imaging quality & 58.19 & \textbf{69.48} \\
\midrule
FFE & Subject consistency & 91.05 & \textbf{96.42} \\
 & Multiple objects & 76.14 & \textbf{82.39} \\
\bottomrule
\end{tabular}
\end{table}

\section{\changedtext{Forward-KL Regularization Details}}
\label{app:forward-kl}

We also test a forward-KL-style distillation regularizer that matches the one-step generator output to the teacher ODE endpoint conditioned on the same noisy latent. The probabilistic objective this regularizer approximates is
\begin{equation}
    \mathcal{L}_{\mathrm{fkl}} =
    \mathbb{E}_{x_{t_0}^{\mathrm{ode}}, c}
    \left[
    D_{\mathrm{KL}}\!\left(
    q_{\mathrm{ODE}}(x_0 \mid x_{t_0}^{\mathrm{ode}}, c)
    \,\|\, p_\theta(x_0 \mid x_{t_0}^{\mathrm{ode}}, t_0, c)
    \right)
    \right],
\end{equation}
where $q_{\mathrm{ODE}}$ is represented by the saved ODE trajectory. Our implementation does not estimate this KL directly. Instead, it optimizes a deterministic squared-error surrogate: the generator prediction is regressed to the stored clean endpoint $x_{\mathrm{tar}}^{\mathrm{ode}}$,
\begin{equation}
    \widehat{\mathcal{L}}_{\mathrm{fkl}}
    =
    \mathbb{E}_{x_{t_0}^{\mathrm{ode}}, x_{\mathrm{tar}}^{\mathrm{ode}}}
    \left\| G_\theta(x_{t_0}^{\mathrm{ode}}, t_0, c) - x_{\mathrm{tar}}^{\mathrm{ode}} \right\|_2^2.
\end{equation}
This squared-error surrogate is added to the DMD and adversarial generator objectives with weight $\lambda_\text{fkl}$. Adding $\lambda_\text{fkl}{=}1$ substantially hurts performance: quality drops to 75.03, total score to 74.83, and dynamic degree to 1.30. Relative to the chunkwise baseline (total 81.60), the total score drops by nearly 7 points and dynamics almost vanish. These results suggest that the deterministic squared-error surrogate for forward-KL regularization is poorly aligned with the distributional objectives used by \method{} in the one-step setting. The objectives therefore conflict, and the extra anchor suppresses motion rather than improving fidelity.

\section{\changedtext{Diversity and Distributional Coverage}}
\label{app:diversity-coverage}

\changedtext{We evaluate whether adding the adversarial objective reduces diversity using two complementary comparisons. First, we measure LPIPS-VGG variation between generated samples for 100 randomly sampled prompts, with four generation seeds per prompt, comparing DMD-only and DMD+GAN at training step 200. Second, we evaluate recall using 256 videos randomly sampled from our dataset. These two protocols are reported together in Table~\ref{tab:diversity-coverage}, but use different evaluation samples.}

\changedtext{LPIPS-VGG is nearly unchanged (0.6437 versus 0.6440), indicating comparable within-prompt perceptual variation. Recall increases from 0.457 to 0.711, providing evidence of improved distributional coverage in this evaluation. These measurements complement the VBench quality gains by assessing within-prompt variation and coverage.}

\begin{table}[ht]
\changedcolor
\centering
\small
\caption{Diversity and coverage with and without the adversarial objective. LPIPS-VGG uses 100 prompts with four generation seeds each; recall uses a separate 256-video evaluation.}
\label{tab:diversity-coverage}
\begin{tabular}{l r r}
\toprule
\textbf{Objective} & \textbf{LPIPS-VGG} & \textbf{Recall} $\uparrow$ \\
\midrule
DMD-only & 0.6437 & 0.457 \\
DMD+GAN & 0.6440 & \textbf{0.711} \\
\bottomrule
\end{tabular}
\end{table}

\section{\changedtext{Four-Step Generation}}
\label{app:four-step-generation}

\changedtext{Although our main focus is one-step generation, the joint DMD and adversarial objective also applies to a larger denoising budget. We train a four-step \method{} model and compare it with four-step Self Forcing on 944 VBench prompts, using Qwen prompt rewriting and the same backbone and inference-step budget. Table~\ref{tab:four-step-additional} shows higher total and quality scores for \method{}, together with improvements in imaging quality and multiple-object composition; the semantic score is slightly lower. The gain at this operating point shows that the benefit of the training objective is not restricted to one-step generation.}

\begin{table}[ht]
\changedcolor
\centering
\small
\setlength{\tabcolsep}{4pt}
\caption{VBench comparison of models trained for four-step generation.}
\label{tab:four-step-additional}
\begin{tabular}{l r r r r r}
\toprule
\textbf{Model} & \textbf{Total} & \textbf{Quality} & \textbf{Semantic} & \textbf{Imaging quality} & \textbf{Multiple objects} \\
\midrule
Self Forcing & 83.46 & 84.77 & \textbf{78.24} & 69.19 & 75.76 \\
\method{} & \textbf{83.85} & \textbf{85.35} & 77.82 & \textbf{71.12} & \textbf{77.74} \\
\bottomrule
\end{tabular}
\end{table}

\section{\changedtext{Inference Latency and Training Cost}}
\label{app:inference-latency}

\changedtext{We benchmark three inference schedules for 21 latent frames over 20 trials on a single H200 GPU. The all-one-step schedule uses one denoising step for every block; FFE uses four steps for the first block and one for each subsequent block; the all-four-step schedule uses four steps for every block. End-to-end timing includes VAE decoding. As shown in Table~\ref{tab:inference-latency}, FFE adds only 0.06\,s to the end-to-end time relative to all-one-step generation, while retaining a similar steady-state block latency. Relative to all-four-step sampling, FFE reduces end-to-end time from 7.72\,s to 4.19\,s, a $1.84\times$ speedup. These measurements complement the FFE quality ablation: the extra denoising budget is confined to the first block rather than repeated throughout the rollout.}

\begin{table}[ht]
\changedcolor
\centering
\small
\caption{Inference timing on one H200 GPU for 21 latent frames and 20 trials. Steady-state block latency excludes the first block; end-to-end time includes VAE decoding.}
\label{tab:inference-latency}
\begin{tabular}{l r r r}
\toprule
\textbf{Schedule} & \textbf{Steady block (ms)} & \textbf{Diffusion (s)} & \textbf{End-to-end (s)} \\
\midrule
All one-step & 117.8 & 2.46 & 4.13 \\
FFE & 120.8 & 2.53 & 4.19 \\
All four-step & 290.9 & 6.07 & 7.72 \\
\bottomrule
\end{tabular}
\end{table}

\paragraph{\changedtext{Training FLOPs.}}
\changedtext{{As detailed in Appendix~\ref{app:implementation-details}, the} reported training configuration uses no R1/R2 gradient penalties. Omitting this regularization reduces training FLOPs by 28.59\%. This saving concerns arithmetic operations during training, not a measured reduction in training time, and is separate from the inference timings in Table~\ref{tab:inference-latency}.}

\end{document}